\documentclass[conference]{IEEEconf}
\IEEEoverridecommandlockouts
\usepackage{cite}
\usepackage{amsmath,amssymb,amsfonts}
\usepackage{optidef}
\usepackage{graphicx}
\usepackage{multirow}
\usepackage{textcomp}
\usepackage{xcolor}
\usepackage{placeins}
\usepackage{float}
\def\BibTeX{{\rm B\kern-.05em{\sc i\kern-.025em b}\kern-.08em
    T\kern-.1667em\lower.7ex\hbox{E}\kern-.125emX}}

\usepackage{mathtools}
\usepackage{wasysym}
\usepackage{bm}
\usepackage{bbm}
\usepackage[font=footnotesize]{subcaption}
\usepackage{booktabs}
\usepackage{array}
\usepackage{dblfloatfix}
\newcolumntype{L}[1]{>{\raggedright\let\newline\\\arraybackslash\hspace{0pt}}m{#1}}
\newcolumntype{C}[1]{>{\centering\let\newline\\\arraybackslash\hspace{0pt}}m{#1}}
\newcolumntype{R}[1]{>{\raggedleft\let\newline\\\arraybackslash\hspace{0pt}}m{#1}}

\usepackage{balance}
\usepackage{graphicx}
\usepackage{multirow}
\usepackage{tabularx}
\usepackage{algpseudocode}
\usepackage[linesnumbered,ruled,vlined,noend]{algorithm2e}
\SetKwRepeat{Do}{do}{while}
\usepackage{hyperref}
\usepackage{xcolor}
\hypersetup{
    colorlinks,
    linkcolor={red!70!black},
    citecolor={blue!70!black},
    urlcolor={blue!90!black}
}
\usepackage{comment}

\makeatletter
\let\oldabs\abs
\def\abs{\@ifstar{\oldabs}{\oldabs*}}
\let\oldnorm\norm
\def\norm{\@ifstar{\oldnorm}{\oldnorm*}}
\makeatother

\title{\LARGE \bf Dynamically Extensible and Retractable Robotic Leg Linkages for Multi-task Execution in Search and Rescue Scenarios}

\author{William Harris$^{\dagger,1}$, Lucas Yager$^{\dagger,1}$, Syler Sylvester$^{2}$, Elizabeth Peiros$^{2}$, Michael C. Yip$^1$, \IEEEmembership{Senior Member, IEEE}%
\thanks{$^\dagger$ Equal Contribution. This work was supported by the U.S. Army's Telemedicine and Advanced Technology Research Center under Project W81XWH-22-C-0089.}%
\thanks{$^1$ W. Harris and L. Yager are with the Mechanical and Aerospace Engineering Department, University of California, San Diego, La Jolla, CA 92093 USA. {\tt\footnotesize \{wharris, lyager\}@ucsd.edu}}%
\thanks{$^2$ E. Peiros, S. Sylvester, and M.C. Yip are with the Electrical and Computer Engineering Department, University of California, San Diego, La Jolla, CA 92093 USA. {\tt\footnotesize \{epeiros, ssylvester, m1yip\}@ucsd.edu}}%
}
    
\begin{document}

\maketitle
\thispagestyle{empty}
\pagestyle{empty}

\begin{abstract}
Search and rescue (SAR) robots are required to quickly traverse terrain and perform high-force rescue tasks, necessitating both terrain adaptability and controlled high-force output. Few platforms exist today for SAR, and fewer still have the ability to cover both tasks of terrain adaptability and high-force output when performing extraction. While legged robots offer significant ability to traverse uneven terrain, they typically are unable to incorporate mechanisms that provide variable high-force outputs, unlike traditional wheel-based drive trains. This work introduces a novel concept for a dynamically extensible and retractable robot leg. Leveraging a dynamically extensible and retractable five-bar linkage design, it allows for mechanically switching between height-advantaged and force-advantaged configurations via a geometric transformation. A testbed evaluated leg performance across linkage geometries and operating modes, with empirical and analytical analyses conducted on stride length, force output, and stability. The results demonstrate that the morphing leg offers a promising path toward SAR robots that can both navigate terrain quickly and perform rescue tasks effectively.
\end{abstract}

\section{Introduction}

Search and rescue (SAR) operations in hazardous environments expose responders to significant physical risk and require extensive training, particularly in hazardous environments such as collapsed structures, disaster zones, or conflict areas, where reaching and assisting victims is both difficult and dangerous. These conditions create a need for systems that can perform physically demanding tasks, reducing responder workload while improving victim outcomes. Currently, SAR teams rely heavily on human personnel and trained dogs, which limits operational speed and reach \cite{delmerico2019current}. Responders are constrained not only by their small numbers and the physical intensity of tasks such as surveying disaster sites, moving rubble, locating trapped individuals, and transporting casualties to safety, but also by the danger of working in unstable, life-threatening environments.

Robotic platforms, including aerial drones and marine vehicles, have been developed to support SAR missions. Such systems require robust mobility, sensing, autonomy, and versatility \cite{mukhopadhyaydesign}, and are optimized for surveying, mapping, or inspection. Most do not physically interact with the environment or transport loads \cite{delmerico2019current, mugorobotics, mukhopadhyaydesign, rubio2019review}, meaning they cannot physically rescue a casualty from danger.  Ground-based robots, while capable of maneuvering rough terrain, often struggle to combine mobility with high load capacity \cite{delmerico2019current, mugorobotics, mukhopadhyaydesign, rubio2019review}.  Some can traverse uneven environments but lack the force needed to extract casualties. An exception is ResQbot, a wheeled casualty extraction robot using a conveyor stretcher \cite{saputra2018resqbot}, which functions only on flat terrain, illustrating the trade-off between locomotion flexibility and payload capacity.

\begin{figure}[!t]
    \centering
    \includegraphics[page=1,width=\linewidth,trim=450 10 400 20,clip]{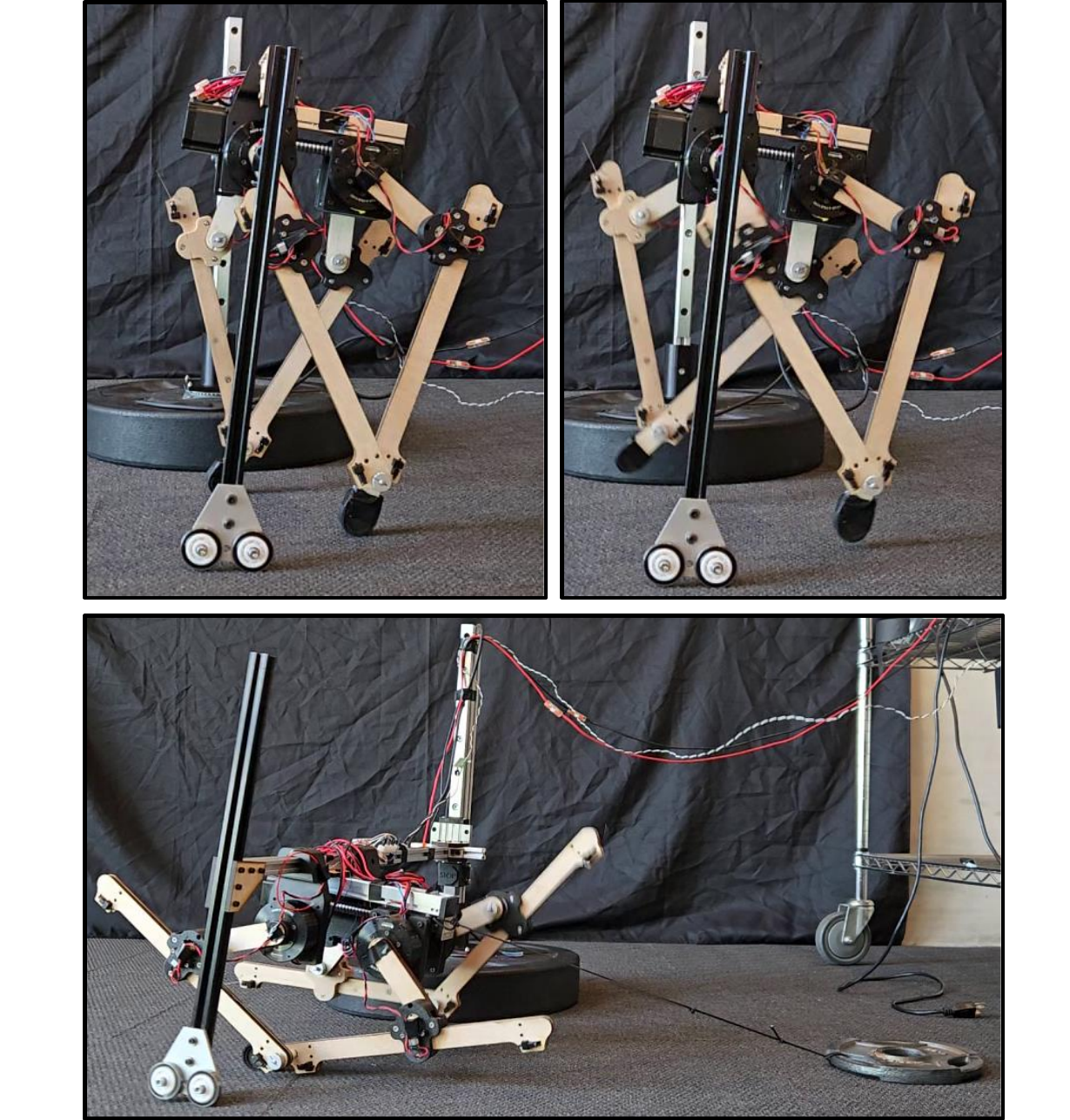}
\caption{Bipedal proof-of-concept SAR robot with adaptable legs switching between 'search' and 'rescue' modes. Upper images show the robot in search mode with legs augmented for fast travel; the lower image shows rescue mode with legs augmented for load-dragging.}
    \label{fig:Biped_2_modes}
\end{figure}

Legged robots offer improved \textit{terrain adaptability}, making them well-suited to navigate uneven terrain. However, this advantage does not address the broader need for \textit{functional adaptability}—the capacity to reconfigure performance for very different tasks, such as rapid traversal versus high-force load pulling. Most legged systems are optimized for mobility and not pulling force \cite{delmerico2019current}, leaving them too specialized to handle the full range of SAR tasks. Recent research has explored morphological transformations to improve versatility, such as adjustable limb lengths \cite{nygaard2021real} that have a modest increase in height (60 cm to 73 cm) or Continuously Variable Transmissions (CVT) to expand the torque–speed range of robotic legs \cite{hur2024continuously}, but has fixed link links and does not expand the reachability of the leg for more challenging terrain. These papers are just initial examples of the increased interest in mechanisms that support the diversity of terrain and torque outputs for legged robots. 

This paper explores a novel approach to addressing the disparity between configurations optimized for rugged locomotion and those for heavy-load operations through a morphing leg architecture. The dual-mode design meets SAR demands for rapid traversal and high-force actions such as dragging casualties or navigating over rubble through changes in mechanical advantage. Exploring the affordances of a morphing leg is nontrivial, so a new modeling and analysis method is introduced for exploring and optimizing leg morphologies along the spectrum of locomotion versus loading. The specific contributions of this work are: (1) the design of a morphing leg capable of switching between speed- and force-optimized modes, (2) methods for modeling and analyzing leg morphologies, and (3) experimental validation through a planar testbed and a bipedal prototype.


\section{Related works}
Research in legged robotics has evolved through single-DOF mechanisms for energy-efficient motion and multi-DOF systems for terrain adaptability. Single-DOF linkages like the eight-bar mechanism~\cite{desai2019analysis} and Chebyshev-based designs~\cite{ju2025development} achieve efficient cyclical gaits through mechanical intelligence with limited adaptability, as shown in hexapod comparisons~\cite{shah2024comprehensive}. While historical analyses~\cite{silva2007historical} document the progression toward multi-DOF platforms, even advanced hexapods~\cite{ma2022design} demonstrate persistent specialization in either speed or force capacity due to fixed morphologies.

Among candidate morphing leg architectures, linkage-based designs provide an attractive compromise between simplicity and versatility. Single-DOF mechanisms such as Jansen’s or Klann’s offer efficient forward motion but lack adaptability, while 4-bar pantographs can generate complex gaits but would require sacrificing efficiency to enable geometric transformation. In contrast, the parallel 5-bar mechanism retains both DOFs for trajectory control, distributes loads evenly across actuators, and allows straightforward augmentation of link lengths to shift mechanical advantage. These characteristics make it a natural foundation for a morphing leg design intended to balance speed and force in SAR contexts.

Five-bar linkage research has progressed along three key dimensions: (1) kinematic studies optimized leg dimensions for quadruped applications \cite{bai2020optimal} and planar manipulators \cite{d2022optimized,kavala2022optimal,sreenivasulu2021inverse}, (2) dynamic implementations achieved amphibious operation \cite{wang2023optimal} and, coaxial actuation \cite{gu2022overconstrained}, and (3) workspace analyses characterized reachable areas \cite{fallahi1994study,hoang2015study}, and singularity-free zones \cite{erwin2024optimal}. Quadruped control strategies \cite{parmar2017conceptualization} and direct-drive comparisons \cite{kenneally2016design} established critical performance baselines, while dyno-kinematic approaches \cite{austin2023dyno} extended dynamic capabilities. Despite these advancements, all implementations fundamentally lack real-time reconfiguration – amphibious designs \cite{wang2023optimal} cannot transition between land/water optimization, coaxial systems \cite{gu2022overconstrained} remain fixed, and high-performance implementations \cite{austin2023dyno} specialize for either speed or force.

This work introduces geometric mode switching to overcome the specialization of SAR robots for either surveying/mapping or victim rescue, using the structure and simplicity of the five-bar linkage.

\begin{figure}[!t]
\vspace{2mm}
\centerline{\includegraphics[width=.8\linewidth, trim=200 125 200 125, clip]{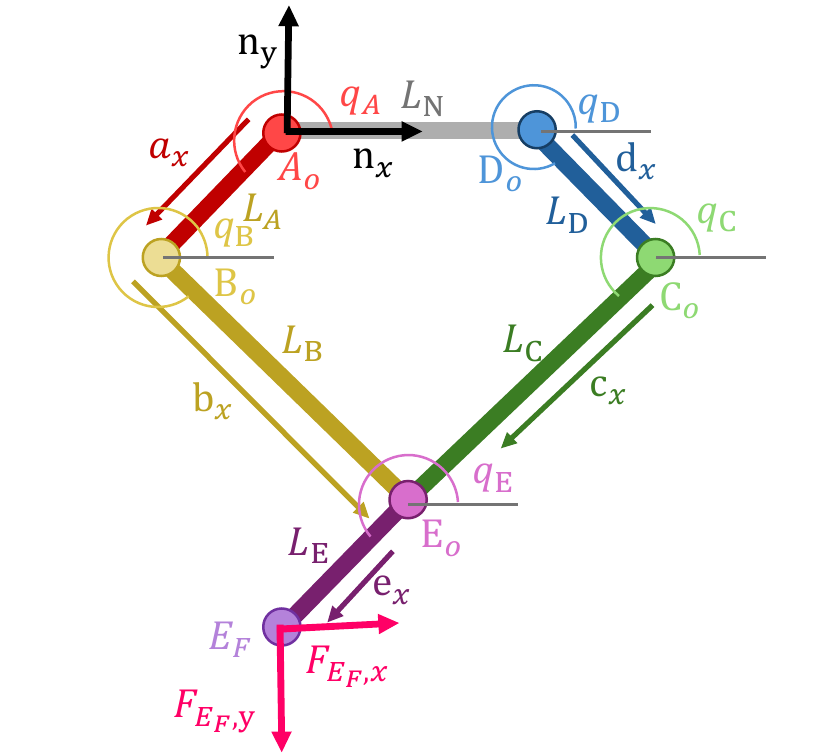}}
\caption{Diagram of 5-bar linkage. The six links are modeled as rigid bodies $\{A, B, C, D, E, N\}$. Links $A$ and $D$ are actively controlled (red and blue). Links $B$ and $C$ are passive (yellow and green). $N$ is the ground link (gray). $E$ (purple) is an ankle extension rigidly attached to link $C$ with the foot at point $E_F$. Each link has an angle $q_{(A, B, C, D, E)}$ measured relative to the world frame with positive sense. All links have an origin point $\{A_o, B_o, C_o, D_o, E_o, N_o\}$ with $\hat{x}$ components along the link length and $\hat{z}$ out of the page.}
\label{fig:linkage_setup}
\end{figure}

\renewcommand{\arraystretch}{1.25}
\begin{table}[!t]
\centering
\caption{List of symbols.}
\resizebox{\columnwidth}{!}{%
\begin{tabular}{|l|c|}  
    \hline
         Quantity & Symbol \\
        \hline
         Length of link $(A, B, C, D, E)$ & $L_{(A, B, C, D, E)}$\\ 
         Distance between $N_o$ and $D_o$ & $L_N$\\
        \hline
        Origin of link $(A, B, C, D, E, N)$ reference frame & $(A, B, C, D, E, N)_o$  \\
        \hline
         Angle from $\hat{n_x}$ to ($\hat{a_x}$,$\hat{b_x}$,$\hat{c_x}$,$\hat{d_x}$,$\hat{e_x}$,$\hat{n_x}$) with $+\hat{n_z}$ sense & $q_{(A,B,C,D,E)}$\\
        \hline
        Torque applied on link $A$ from $N$ with $+\hat{n_z}$ sense & $T_A$\\
        Torque applied on link $D$ from $N$ with $+\hat{n_z}$ sense & $T_D$\\
        \hline
        Horizontal ground reaction force on the foot & $R^F_x$\\
        Vertical ground reaction force on the foot & $R^F_y$\\
        \hline
        Horizontal force from the foot on the ground & $F_{E_F,x}$\\
        Vertical force from the foot on the ground & $F_{{E_F},y}$\\
        \hline
        Cartesian coordinates of the foot, $E_F$& $x_{E_F},y_{E_F}$\\
        Cartesian coordinates of where links $B$ and $C$ meet, $E_o$.& $x_{E_o},y_{E_o}$\\
    \hline
\end{tabular}
}
\label{tab:symbols}
\end{table}
\renewcommand{\arraystretch}{1.0} 

\section{Methods}

A parallel 5-bar mechanism was chosen as the adaptive leg structure. Its two degrees of freedom define a planar foot path while distributing load evenly between actuators, and its simple geometry supports link-length augmentation for mode switching. To determine which links to vary, a combination of workspace evaluation, kinematic modeling, and force analysis was used to compare candidate geometries. The resulting design was then implemented and evaluated on a testbed under static conditions.

\subsection{Linkage Model}
\label{sec: model}

 The geometry of the 5-bar linkage system was set according to the diagram shown in Figure \ref{fig:linkage_setup}. The six links, including a possible foot link, are modeled as rigid bodies $\{A, B, C, D, E, N\}$, and links $A, B, C, D,$ and $N$ form a closed kinematic chain. Links $A$ and $D$ are the actively controlled links, links $B$ and $C$ are the passive links, $N$ is the ground link, and $E$ is an ankle extension link. This link is rigidly attached to link $C$ with the foot attached to the end at a point $E_F$ such that $q_E=q_C$. Each link has an angle $q_{(A, B, C, D, E)}$ measured relative to the world frame, $N$, with a positive sense. All links have an origin point on the link $\{A_o, B_o, C_o, D_o, E_o, N_o\}$ with their components $\hat{x}$ directed along the length of the link and $\hat{z}$ directed out of the page. $A_o$ is positioned at $(0,0)$ and $D_o$ is positioned at $(L_N,0)$. The two motors are mounted on the ground link at points $A_o$ and $D_o$ and apply torques $T_A$ and $T_D$ to links $A$ and $D$, at $A_o$ and $D_o$, respectively.

\subsection{Testbed}

\begin{figure}[!t]
    \vspace{2mm}
    \centering
    \includegraphics[width=0.85\linewidth, trim=80 70 130 60, clip]{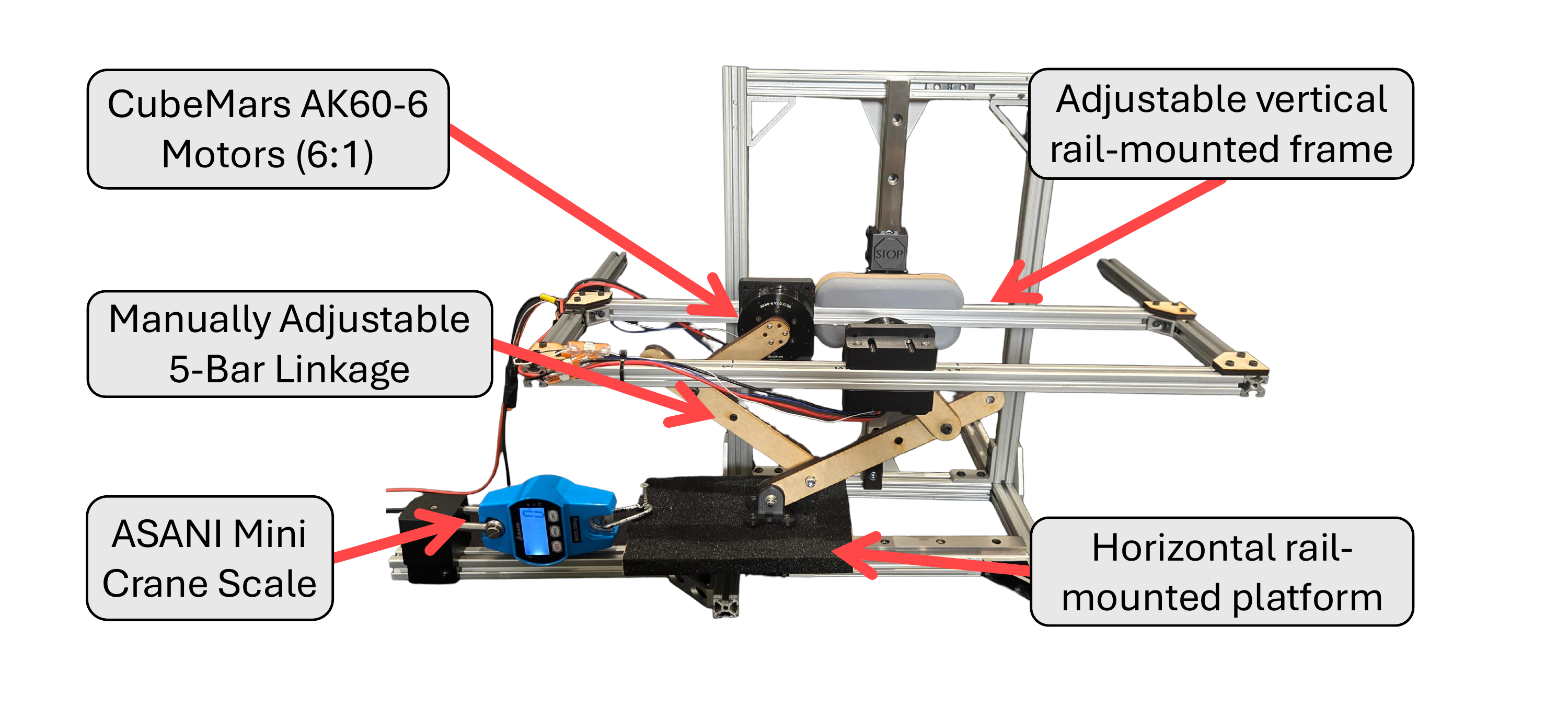}
    \caption{Testbed setup with vertical rail-mounted leg, horizontal force measurement platform, and three-motor actuation system for evaluating adaptive linkage performance.}
    \label{fig:testbed}
\end{figure}

Figure \ref{fig:testbed} shows the custom testbed constructed to evaluate the adaptive 5-bar linkage leg under static conditions. The testbed consisted of a lockable vertical linear rail where the leg was mounted. The leg can be lifted and repositioned during walking trials or clamped down to estimate maximum ground reaction forces. Ground contact was replicated using a horizontal linear rail with a pillow-block attachment for securing the foot to simulate pushing without slipping. An ASANI Mini Crane Scale with a range of 300kg was mounted to measure the horizontal pulling forces at the foot. 

The testbed leg was fabricated using laser-cut plywood with additional holes added to the $B$ and $C$ links to allow for testing of different link lengths. Rotation of the $A$ and $D$ links was achieved with Cubemars AK60-6 V3.0 BLDC motors, chosen for their nominal torque capacity of 3 Nm. 

 

\subsection{Dynamic Leg}

\begin{figure}[t!]
    \centering
    \includegraphics[width=0.85\linewidth, trim=15 45 15 75, clip]{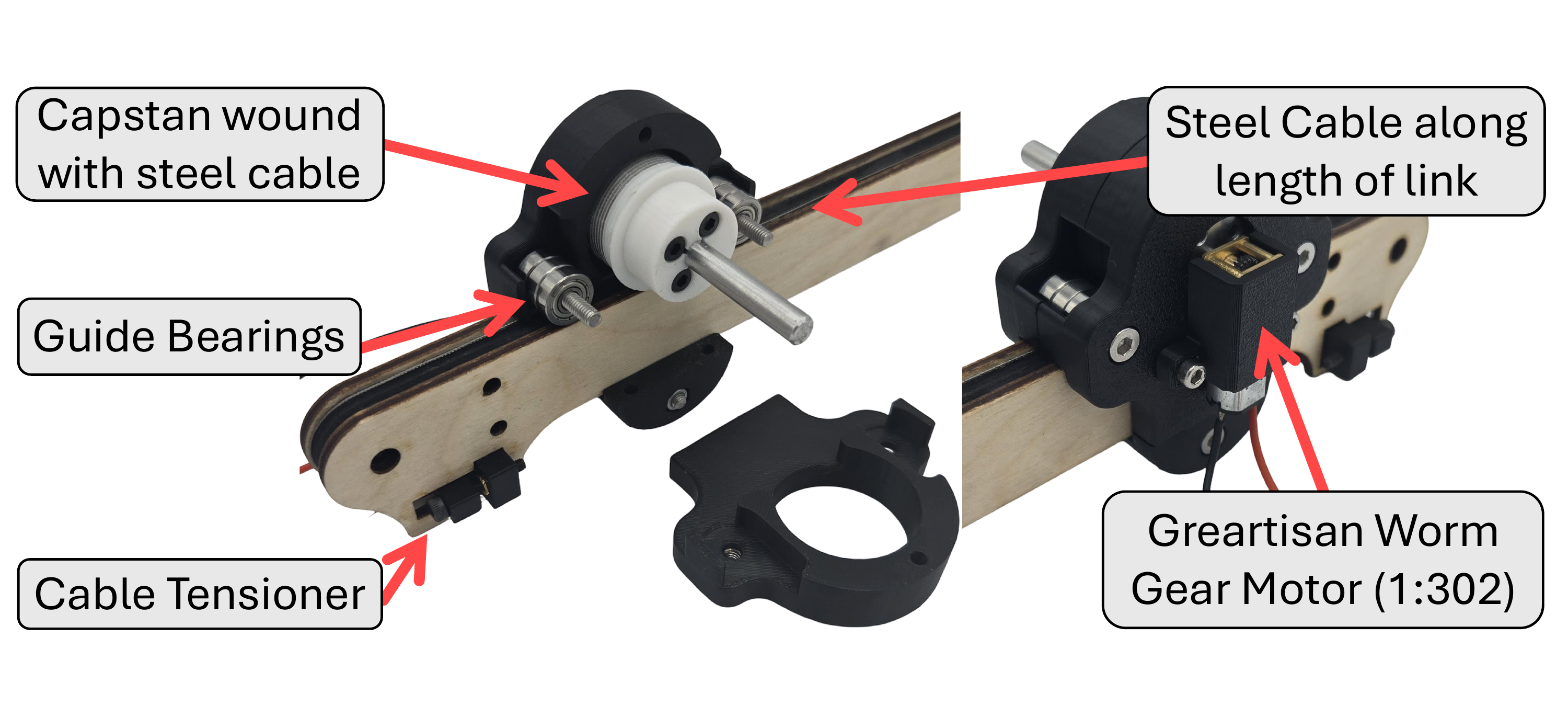}
    \caption{Capstan-driven actuator mechanism showing motor-driven spool system for varying effective link length through cable tension adjustment. The overall length change is 64:273 mm.}
    \label{fig:linear link actuator}
\end{figure}

To achieve a dynamically length-changing 5-bar linkage, a combination of capstan drives and a linear stage with a stepper motor was used shown in Figures \ref{fig:Biped_2_modes} and \ref{fig:transformation}. The linear stage has an effective travel length of 200 mm with a Nema 23 stepper motor that actuates a ball screw to slide the carriage along a linear rail. The linear stage was used as the ground link $N$ to which motors were attached. A capstan drive changes the lengths of the $B$ and $C$ links. The capstan spools are attached at the joints between links A–B and D–C, seen in Figure \ref{fig:linear link actuator}. The capstan is driven by a Greartisan 12 V DC worm gear micro-motor with a 1:302 gear reduction. A steel cable wound around each spool was routed along the outer edge of the link and held in tension. As the spool winds or unwinds, the cable shifts its contact point and thereby alters the effective length of the link. To maintain smooth and constrained motion, the spool was supported by a coupler with three bearings, which kept it tangent to the link while minimizing drag. This configuration enabled reliable linear actuation of the passive links in a compact package, supporting repeatable transitions between configurations without problems of debris jamming up lead screw designs, or the mechanical instability scissor-lifts have.

\begin{figure}[!t]
\vspace{2mm}
 \centerline{\includegraphics[page=1,width=\linewidth, trim=80 180 80 170, clip]{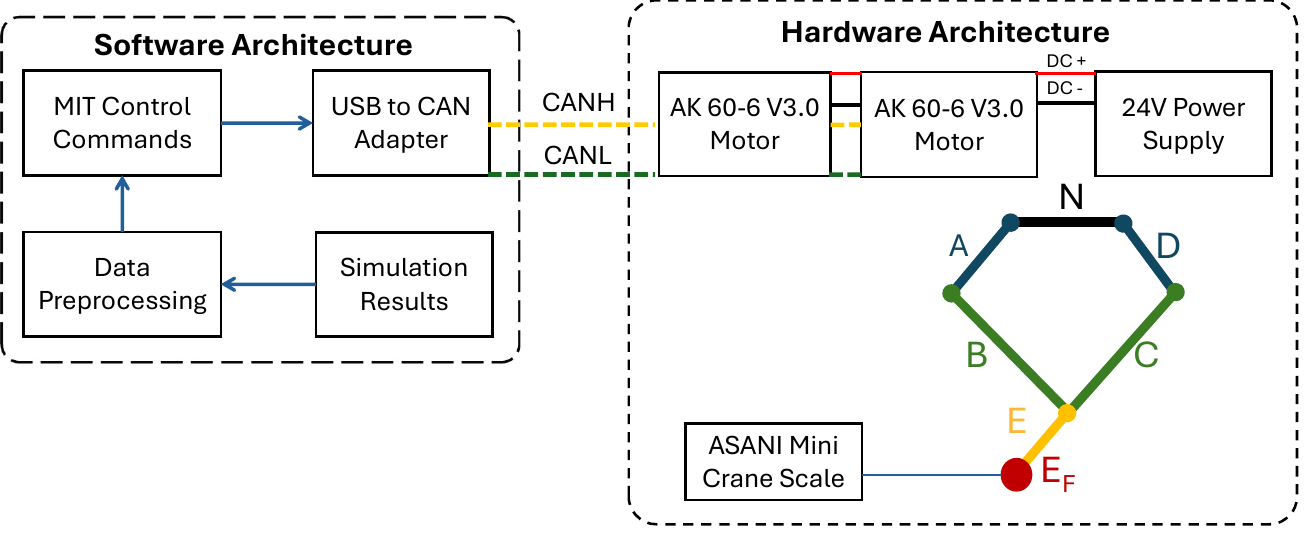}}
\caption{System architecture for the robotic leg testbed. Two AK60-6 V3.0 motors are simultaneously controlled by a remote desktop over CAN bus, shown in green and yellow dashed lines. The remote desktop converts simulated position, velocity, and torque metrics into MiT commands sent to the motors via a USB to CAN converter. The crane scale is used to record the pulling force of the leg to validate simulation results.} \end{figure}

\begin{figure*}[!t]
\vspace{2mm}
 \centerline{\includegraphics[page=2,width=\linewidth, trim=5 245 15 170, clip]{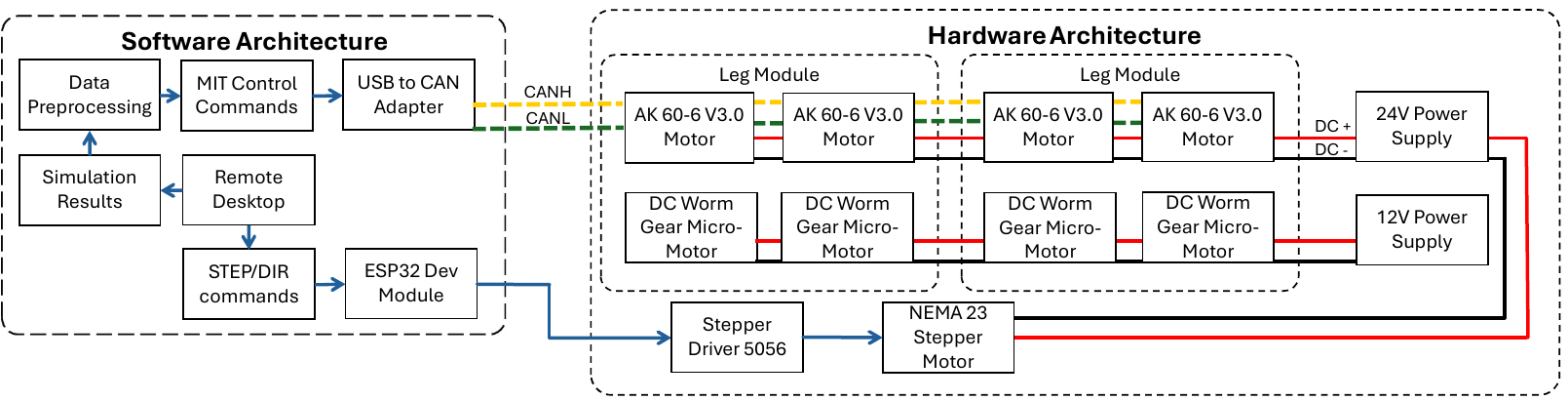}}
\caption{Full system architecture for the bipedal system. Four AK60-6 V3.0 motors are simultaneously controlled over CAN, shown in green and yellow dashed lines, using simulated position and velocity metrics to recreate simulated footpaths. The ESP32 Dev Module and stepper driver actuate a NEMA 23 stepper motor to manipulate the length of the ground length $N$, distance between each motor, through the STEP/DIR protocol. A 24 V power supply is used for the AK60-6 V3.0 motors and the stepper motor. A separate 12 V power supply powers individual DC worm gear micro-motors to shorten and lengthen the $B$ and $C$ links. All power lines are shown in red and ground shown in black} \end{figure*}
\vspace{-2mm}
\subsection{Forward Kinematics}

\begin{figure}[t!]
    \centering
    \begin{subfigure}[b]{\linewidth}
    \centering
    \includegraphics[width=0.70\linewidth, trim=10 80 50 20, clip]{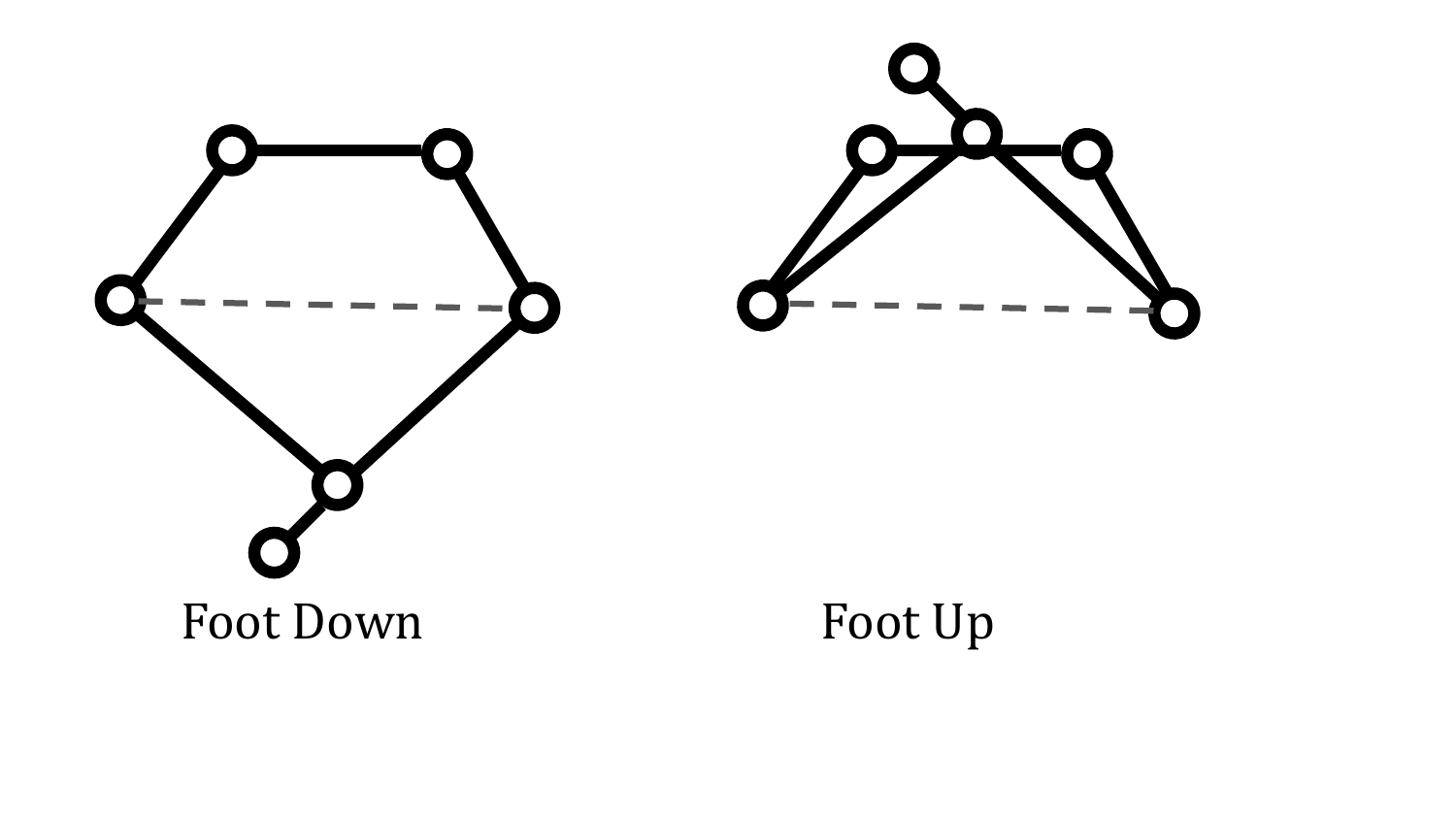}
    \caption{The two forward kinematics solutions for a given $q_A$ and $q_D$.}
    \label{fig:footupdown}
    \end{subfigure}
    \par
    \bigskip
    \begin{subfigure}[b]{\linewidth}
    \centering
    \includegraphics[width=.8\linewidth, trim=0 200 0 10, clip]{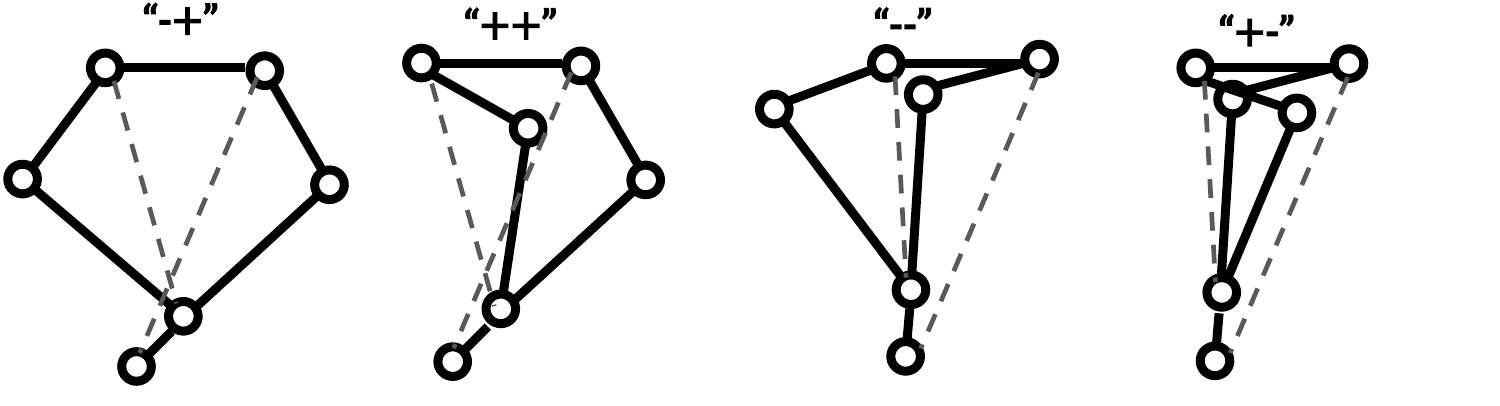}
    \caption{The four possible inverse kinematics solutions for a given foot position $(x_{E_F},y_{E_F})$. The ''-+" inverse solution will be used for the simulation of the leg.}
    \label{fig:workingmodes}
    \end{subfigure}
    \caption{Forward and inverse kinematic solutions for position.}
\end{figure}

Forward kinematic equations can be used to find the position of the foot of the leg in terms of the angles of the actuated links $q_A$ and $q_D$. For any $q_A$ and $q_D$ where there is a solution, there are two solutions for where the foot can be an `up' configuration and a `down' configuration\cite{hoang2015study}, which can be seen in Figure \ref{fig:footupdown} as the $B$ and $C$ links can be reflected over the line between $B_o$ and $C_o$.

 Based on Figure \ref{fig:linkage_setup}, the coordinates of the point $E_F$ are:
 \begin{equation}
      \begin{aligned}
        \label{eq:xF CD side position}
        x_{E_F}=L_N+(L_C+L_E)\cos(q_C)+L_D\cos(q_D)\\
        y_{E_F}=(L_C+L_E)\sin(q_C)+L_D\sin(q_D)
 \end{aligned}
 \end{equation}

Links $A, B, C, D,$ and $N$ form a closed kinematic chain; therefore, a geometric constraint is applied to eliminate $q_B$ and $q_C$ for the equations:
\begin{equation}
\label{eq:position loop}
    \begin{aligned}
        L_A\cos(q_A)+L_B&\cos(q_B)=\\&L_C\cos(q_C)+L_D\cos(q_D)+L_N\\
        L_A\sin(q_A)+L_B&\sin(q_B)=\\&L_C\sin(q_C)+L_D\sin(q_D)
    \end{aligned}
\end{equation}

Equations for $x_{E_F}$ and $y_{E_F}$ (the coordinates of the foot) can then be found as functions of $q_A$ and $q_D$:
\begin{equation}
        \begin{aligned}
                x_{E_F}=&L_N+L_D\cos(q_D)+\\&(L_C+L_E)\cos(\varphi\pm \arccos (\frac{L_C^2+G-L_B^2}{2L_C\sqrt{G}}))\\
                y_{E_F}=&L_D\sin(q_D)+\\&(L_C+L_E)\sin(\varphi\pm \arccos (\frac{L_C^2+G-L_B^2}{2L_C\sqrt{G}}))
        \end{aligned}
\end{equation}
where
\begin{align*}
    G=&L_A^2+L_D^2+L_N^2-2L_AL_D\cos(q_A-q_D)\\&-2L_AL_N\cos(q_A)+2L_DL_N\cos(q_D)
\end{align*} 
and
\begin{align*}
    \varphi=\operatorname{atan2}(L_A\sin(q_A)-L_D\sin(q_D),\\L_A\cos(q_A)-L_N-L_D\cos(q_D))
\end{align*} 
To solve for the velocity equations, the derivatives of Equations \ref{eq:xF CD side position} and \ref{eq:position loop} are used:
\begin{equation}
        \begin{aligned}
            \dot{x_{E_F}}=-(L_C+L_E)\dot{q_C}\sin(q_C)-L_D\dot{q_D}\sin(q_D)\\
            \dot{y_{E_F}}=(L_C+L_E)\dot{q_C}\cos(q_C)+L_D\dot{q_D}\cos(q_D)
        \end{aligned}
\end{equation}
\begin{equation}
        \begin{aligned}
            L_A\dot{q_A}\sin(q_A)+L_B&\dot{q_B}\sin(q_B)=\\&L_C\dot{q_C}\sin(q_C)+L_D\dot{q_D}\sin(q_D)\\
            L_A\dot{q_A}\cos(q_A)+L_B&\dot{q_A}\cos(q_B)=\\&L_C\dot{q_A}\cos(q_C)+L_D\dot{q_A}\cos(q_D)
        \end{aligned}
\end{equation}

Since $\dot{q_B}$ and $\dot{q_C}$ are dependent on $\dot{q_A}$ and $\dot{q_D}$: 
\begin{equation}
\label{eq:kin_jacobian}
    \begin{bmatrix}
        \dot{x_{E_F}}\\ \dot{y_{E_F}}
    \end{bmatrix}
    =
    \begin{bmatrix}
        J_{11} & J_{12}\\ J_{21} & J_{22}
    \end{bmatrix}
    \begin{bmatrix}
        \dot{q_A}\\ \dot{q_D}
    \end{bmatrix}
\end{equation}
where $J_{11}$, $J_{12}$, $J_{21}$, and $J_{22}$ make up the Jacobian:
\begin{align*}
\begin{cases}
    J_{11}=\frac{L_A(L_C+L_E)\sin(q_A-q_B)\sin(q_C)}{L_C\sin(q_B-q_C)}\\
    J_{12}=\frac{L_D(L_C+L_E)\sin(q_B-q_D)\sin(q_C)}{L_C\sin(q_B-q_C)}-L_D\sin(q_D)\\
    J_{21}=\frac{-L_A(L_C+L_E)\sin(q_A-q_B)\cos(q_C)}{L_C\sin(q_B-q_C)}\\
    J_{22}=\frac{-L_D(L_C+L_E)\sin(q_B-q_D)\cos(q_C)}{L_C\sin(q_B-q_C)}+L_D\cos(q_D)
\end{cases}
\end{align*}

\subsection{Inverse Kinematics}
To find the position and angular velocities for the foot to move along a given foot path with specified coordinates and velocities, the inverse kinematics equations were derived. For any accessible foot position, there are up to 4 possible configurations of the linkage that could reach that foot position, as shown in Figure \ref{fig:workingmodes}. This is because links $C$ and $D$ can be reflected on the line between the foot $E_F$ and the point $D_o$, and links $A$ and $B$ can be reflected over the line between point $A_o$ and point $E_o$. Each of these configurations is a different 'working mode', and to switch between them, the linkage would need to pass through a singularity\cite{d2022optimized}. To allow for $L_N=0, L_A=L_D$ and $L_B=L_C$ without the linkage being in a singularity, the `-+' is used as the working mode for all simulations.
The possible solutions of $q_D$ for a specified foot position $(x_{E_F},y_{E_F})$ are:

\begin{multline}
       q_D=\operatorname{arctan2}(y_{E_F},x_{E_F}-L_N)\pm\\\arccos{\frac{L_C^2-L_D^2-y_{E_F}^2-(L_N-x_{E_F})^2}{-2L_D\sqrt{y_{E_F}^2+(L_N-x_{E_F})^2}}}
\end{multline}
Once the possible values for $q_D$ are known, $x_{E_o}$ and $y_{E_o}$ can be found as:
\begin{equation}
    x_{E_o}=\frac{L_Cx_{E_F}+L_E(L_N+L_D\cos(q_D))}{L_C+L_E}
\end{equation}
\begin{equation}
    y_{E_o}=\frac{L_Cy_{E_F}+L_EL_D\sin(q_D)}{L_C+L_E}
\end{equation}
The possible values of $q_A$ are then:
\begin{multline}
    q_A=\operatorname{arctan2}(y_{E_o},x_{E_o})\pm\\
    \arccos{\frac{L_B^2-L_A^2-x_{E_o}^2-y_{E_o}^2}{-2L_A\sqrt{x_{E_o}^2+y_{E_o}^2}}}
\end{multline}
Once the angular positions are known, the Jacobian derived in Equation \ref{eq:kin_jacobian} can be used to find $\dot{q_A}$ and $\dot{q_D}$ according to the relationship:
\begin{equation}
    \begin{bmatrix}
        \dot{q_A}\\ \dot{q_D}
    \end{bmatrix}
    =
    {J(q)}^{-1}
    \begin{bmatrix}
        \dot{x_{E_F}}\\ \dot{y_{E_F}}
    \end{bmatrix}
\end{equation}
\renewcommand{\arraystretch}{1.25}
\begin{table}[!t]
\vspace{2mm}
\centering
\caption{Link Length Ranges}
\resizebox{\columnwidth}{!}{%
\begin{tabular}{|l|c|c|c|c|}  
    \hline
         &$L_A,L_D$ & $L_B,L_C$ & $L_N$ & $L_E$ \\
        \hline
        Base Length (cm)&10&20&10&5\\
        \hline
        Retracted Length (cm)&5&15&0&0\\
        \hline
        Extended Length (cm)&15&30&20&10\\
        \hline
\end{tabular}}
\label{tab:link_lengths}
\end{table}
\renewcommand{\arraystretch}{1}
\subsection{Evaluating Linkage Performance}
For the baseline configuration to be analyzed, appropriate lengths for each link were chosen based on previous 5-bar robot leg designs \cite{d2022optimized}\cite{kenneally2016design}. The effect of changing the length of the symmetric pairs of links, $A$ and $D$, and $B$ and $C$, the ground link $N$, and the ankle link $E$ was compared. The extended and retracted lengths that were analyzed are shown in Table \ref{tab:link_lengths}. To decide which link(s) would be most useful to change in length, the horizontal forces generated at the foot were found throughout the usable workspace.

\begin{figure}[t!]
\vspace{2mm}
\centerline{\includegraphics[page=1,width=\linewidth,trim=50 190 80 0, clip]{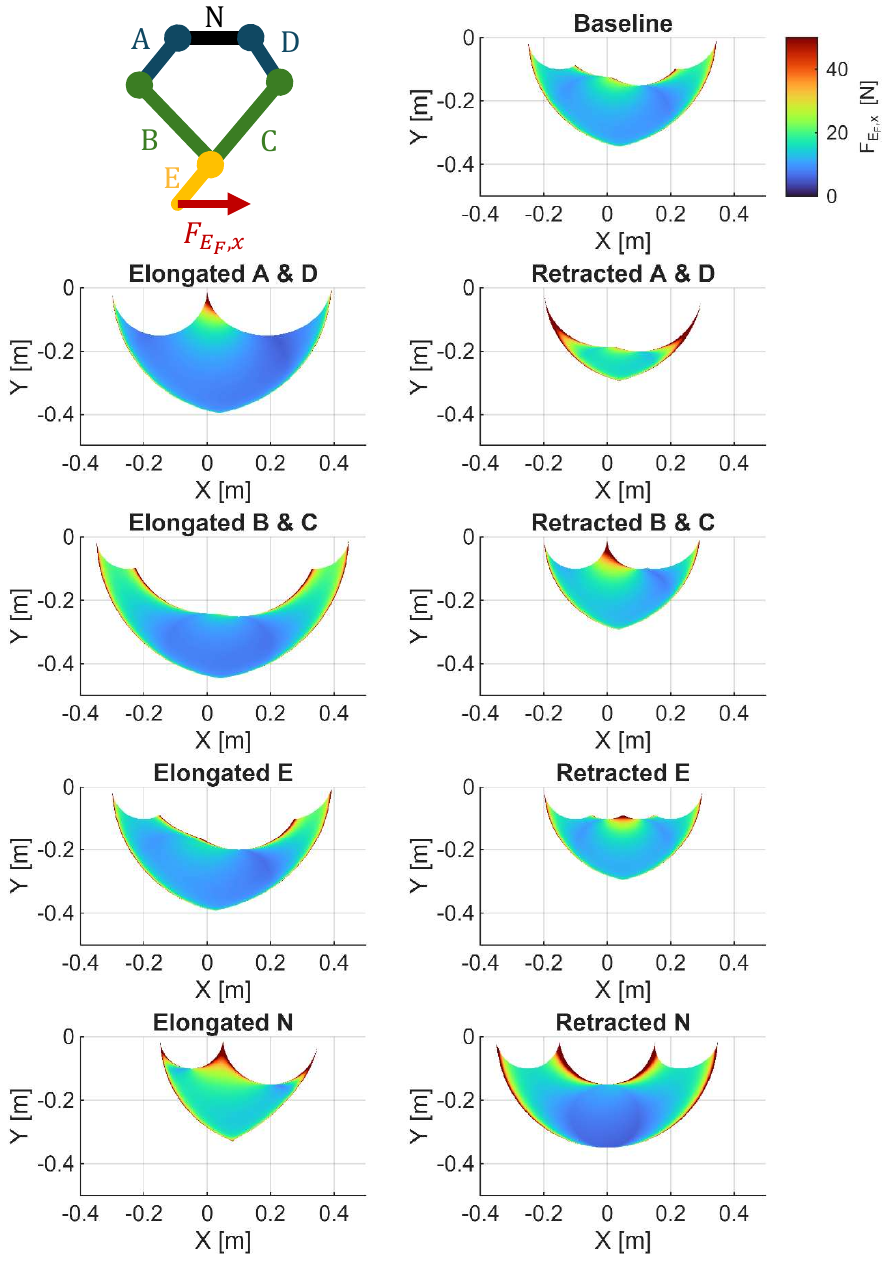}}
\caption{Effects of changing different linkage lengths on theoretical max horizontal pushing force and workspace area. Link lengths are shown in Table \ref{tab:link_lengths}. Decreasing $L_A$ and $L_D$ increases forces throughout the workspace, especially in upper corners, while decreasing workspace size. Increasing $L_B$ and $L_C$ lowers central pushing forces while increasing workspace size. Increasing $L_E$ lowers the pushing forces around the workspace center. Increasing $L_N$ increases the central pushing force and decreases the workspace size.}
\label{fig:change_comparison}
\end{figure}

\begin{figure}
\vspace{2mm}
\centerline{\includegraphics[page=2,width=\linewidth,trim=50 190 80 0, clip]{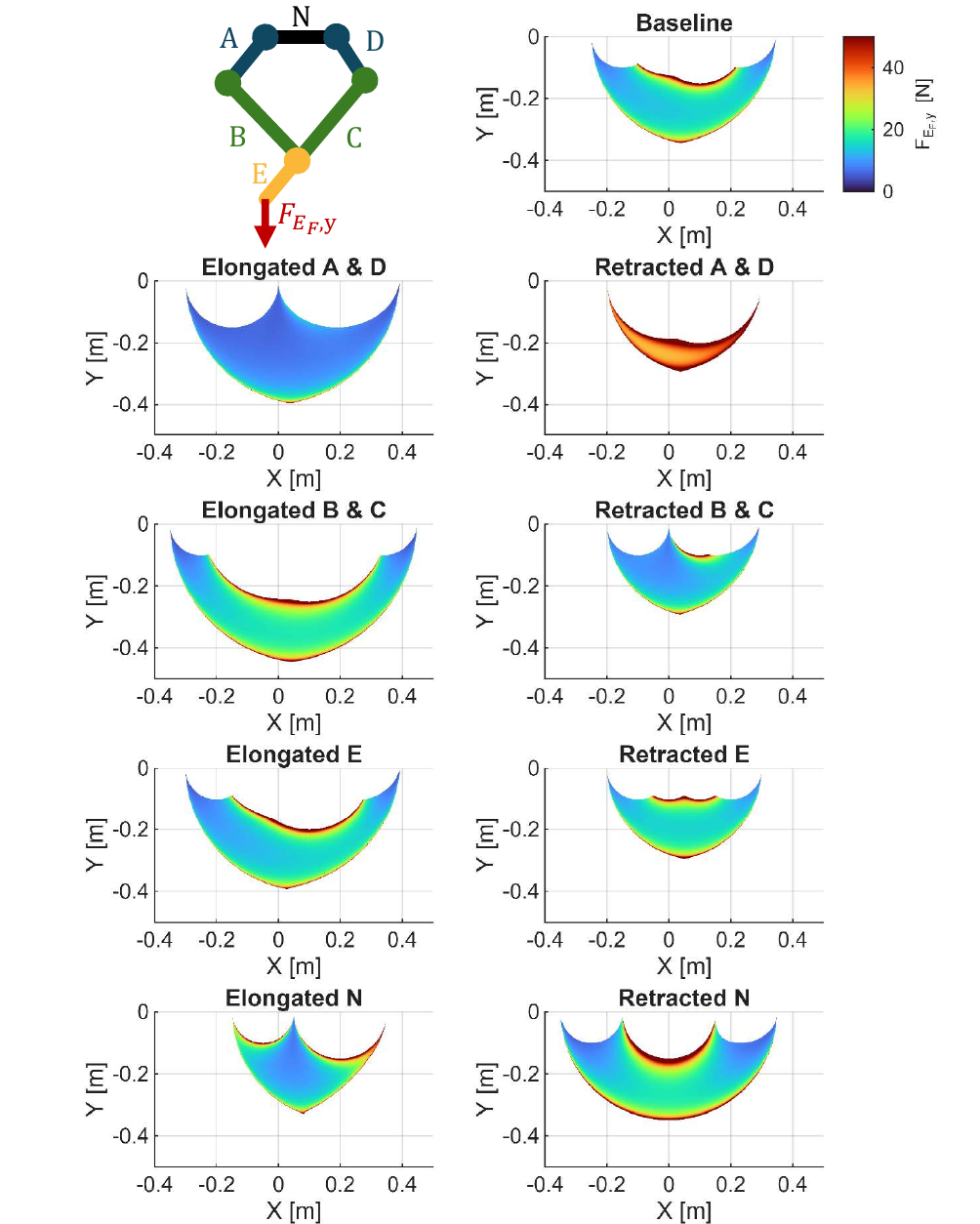}}
\caption{Effects of changing different linkage lengths on theoretical max vertical lifting force and workspace area. Link lengths are shown in Table \ref{tab:link_lengths}. Decreasing $L_A$ and $L_D$ increases forces throughout the workspace. Increasing $L_B$ and $L_C$ increases the lifting forces. Changing $L_E$ has little impact on vertical forces. Increasing $L_N$ decreases the lifting forces achievable in the center of the workspace.}
\label{fig:vert_forces}
\end{figure}

\subsubsection{Finding Workspace}
The total area reachable by the foot is defined by the positions where $q_A=q_B$ or $q_A=q_B+\pi$ and $q_D=q_C$ or $q_D=q_C + \pi$ \cite{fallahi1994study}. These workspace boundaries are singularities of the linkage. The full workspace may include points where the `elbows' $B_o$ and $C_o$ are below the position of the foot, $E_F$, when the ground is taken to be parallel to the ground link $N$ the foot would not be in contact with the ground when the elbows are below it, so these points will be filtered out of the workspace to find the usable workspace. 
\subsubsection{Finding Static Forces}
The theoretical static pushing forces were found using the Jacobian of the 5-bar system, defined by: 
\begin{equation}
\label{eq:jacobian_F_tau}
    \begin{bmatrix}
    F_{E_F,x} \\ F_{E_F,y}
    \end{bmatrix}
    = {(J(q)^T)}^{-1}
    \begin{bmatrix}
    \tau_A \\ \tau_D
    \end{bmatrix}
\end{equation}
following the relationship $\tau=J(q)^TF$\cite{craig1979systematic}, and maximizing the forces achievable at the foot with the magnitude motor torque limited to 1 Nm, so that the forces simulated are newtons per Nm of torque of the motors.

\subsubsection{Condition Number}
The condition number $\kappa$ is the ratio of the largest singular value to the smallest singular value of a matrix. The condition number can be used to determine the error amplification between the input and output and to determine when the linkage is close to a singularity \cite{merlet2006jacobian}. The condition number of the Jacobian is shown in Equation \ref{Eq:Cond_Num}:

\begin{equation}
    \kappa(J(q))=||J(q)||||J^{-1}(q)||
    \label{Eq:Cond_Num}
\end{equation}

\section{Experiments and Simulations}

To evaluate the effectiveness of a morphing five-bar linkage leg, both simulation studies and real-world experiments were conducted. The simulations were used to identify which link length variations most effectively supported the two target modes: “search,” favoring large workspace and obstacle clearance, and “rescue,” prioritizing concentrated horizontal high-force. To validate the simulation results, the horizontal pushing force produced by the selected link lengths was measured using a physical testbed. Finally, to test the proposed methods for linkage actuation and walking characteristics of the leg, a bipedal prototype on a boom arm was built.

\subsection{Simulated Length Changes}

Simulations of the horizontal pushing force on the foot were conducted to determine which link length variations best supported the desired ``search'' and ``rescue'' modes.

In the ``search'' mode, the leg requires a large workspace to enable obstacle clearance and a lower $y$-domain to raise the chassis over mud or shallow water. In the ``rescue'' mode, the leg must generate a concentrated region of high pushing force, centered near the ground link and away from workspace boundaries, to allow stable and effective dragging. To identify the configurations that best enable transitions between these two requirements, simulations of the maximum horizontal pushing force and horizontal lifting force were performed for varying link lengths. The simulation results are summarized in Figure \ref{fig:change_comparison} and \ref{fig:vert_forces} and Table \ref{tab:length_change_ratings}. 

\begin{figure}[!t]
    \vspace{2mm}
    \centering
    \includegraphics[width=.8\linewidth, trim=35 35 60 30, clip]{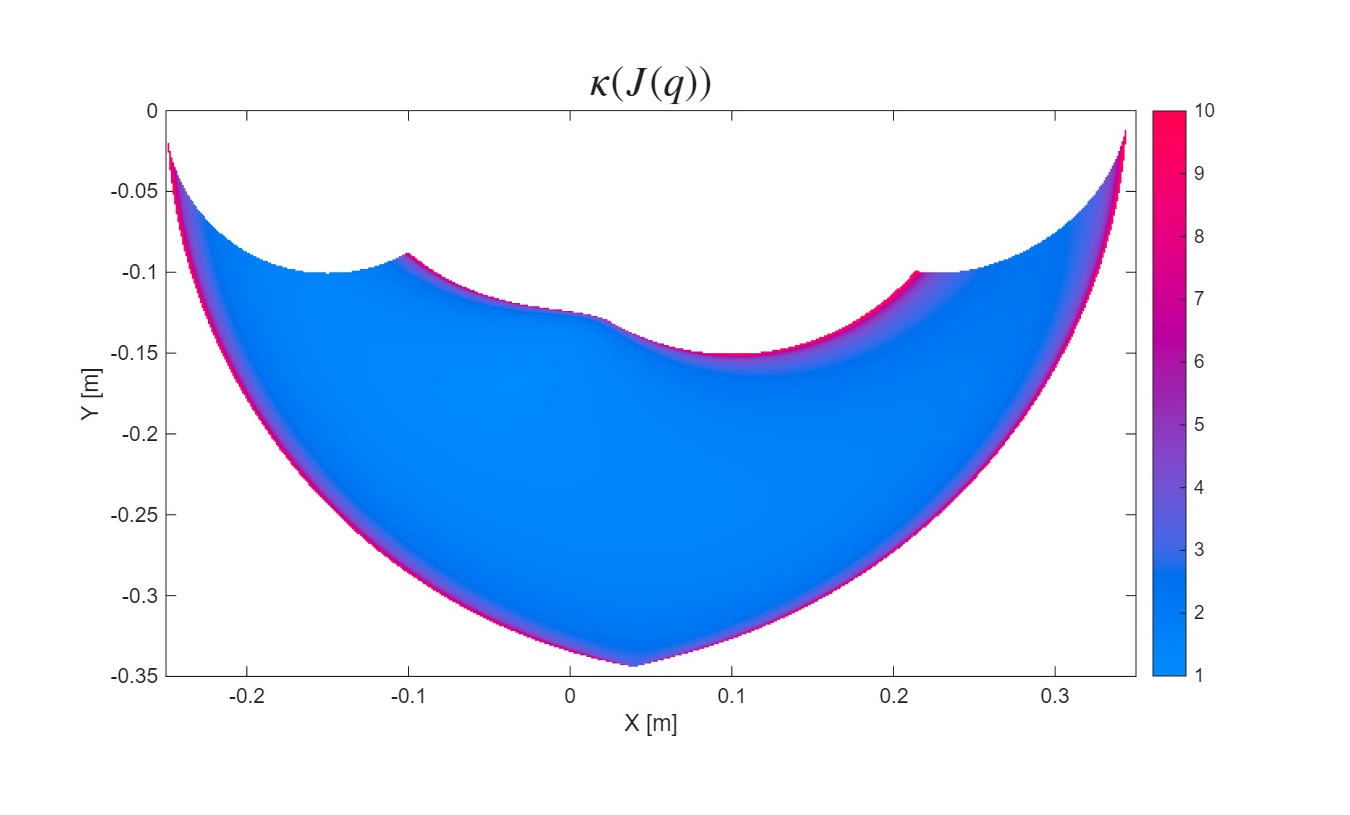}
    \caption{Condition number, $\kappa(J(q))$ throughout the workspace of the baseline configuration. A higher condition number indicates that the position is closer to a singularity and should be avoided.}
    \label{fig:condition_num}
\end{figure}

Elongating the controlled links, $A$ and $D$, increased workspace size and maximum chassis height, with a narrow central horizontal high-force region, and lower maximum vertical forces throughout the workspace. Retracting them led to a small workspace with higher horizontal forces throughout, with maxima near the left and right boundaries, but reduced overall reach. With the links retracted, high vertical forces are present throughout the whole workspace. 

\renewcommand{\arraystretch}{1.25}
\begin{table}[!t]
    \centering
    \caption{Summary of Workspace Sizes}
    \resizebox{.8\columnwidth}{!}{%
    \begin{tabular}{|c|c|}
    \hline
         &Workspace Size $(m^2)$\\
         \hline
        Baseline & .0902\\
        \hline
        Retracted $A$ and $D$& .0291\\
        Elongated $A$ and $D$& .1247 \\
        \hline
        Retracted $B$ and $C$& .0667 \\
        Elongated $B$ and $C$& .1336 \\
        \hline
        Retracted $E$ &.0671 \\
        Elongated $E$ &.1124 \\
        \hline
        Retracted $N$ &.1251 \\
        Elongated $N$ &.0639 \\
        \hline
    \end{tabular}
    }
    \label{tab:length_change_ratings}
\end{table}
\renewcommand{\arraystretch}{1}

Changes to the passive links $B$ and $C$ significantly affected both workspace and force distribution. Elongation created a much larger workspace, with high-force regions near the upper corners. Retraction, by contrast, yielded higher forces throughout most of the workspace and produced a large, centered region of moderate-to-high pushing force well suited for dragging. However this decreased the maximum vertical force achievable. 

Elongation of the ankle link $E$, expanded the reachable area but lowered the pushing force, though a strong region remained in the top right. Setting $L_E=0$  created a concentrated high horizontal force zone in the workspace center. Changes in the length of $E$ had little effect on the maximum vertical forces throughout the workspace.

The ground link $N$ had a particularly strong effect on horizontal force; elongation increased central pushing force while shrinking workspace size, whereas retraction enlarged the workspace with force regions shifted toward the upper edges. These trends highlight $N$ as a key link for tuning between modes. 

Based on these results, it was decided that the ground link $N$, and passive links $B$ and $C$ should be linearly actuated to achieve the desired conformation between the 'search' and 'rescue' modes. While the highest forces were generated when $L_E=0$, this would put the joint between $B$ and $C$ on the ground, where it would be more likely to fill with debris. For this reason, it was decided that the length of $E$ should be minimal but not 0.

\begin{figure*}[!t]
\vspace{2mm}
\centerline{\includegraphics[width=.8\linewidth,trim=75 30 40 20, clip]{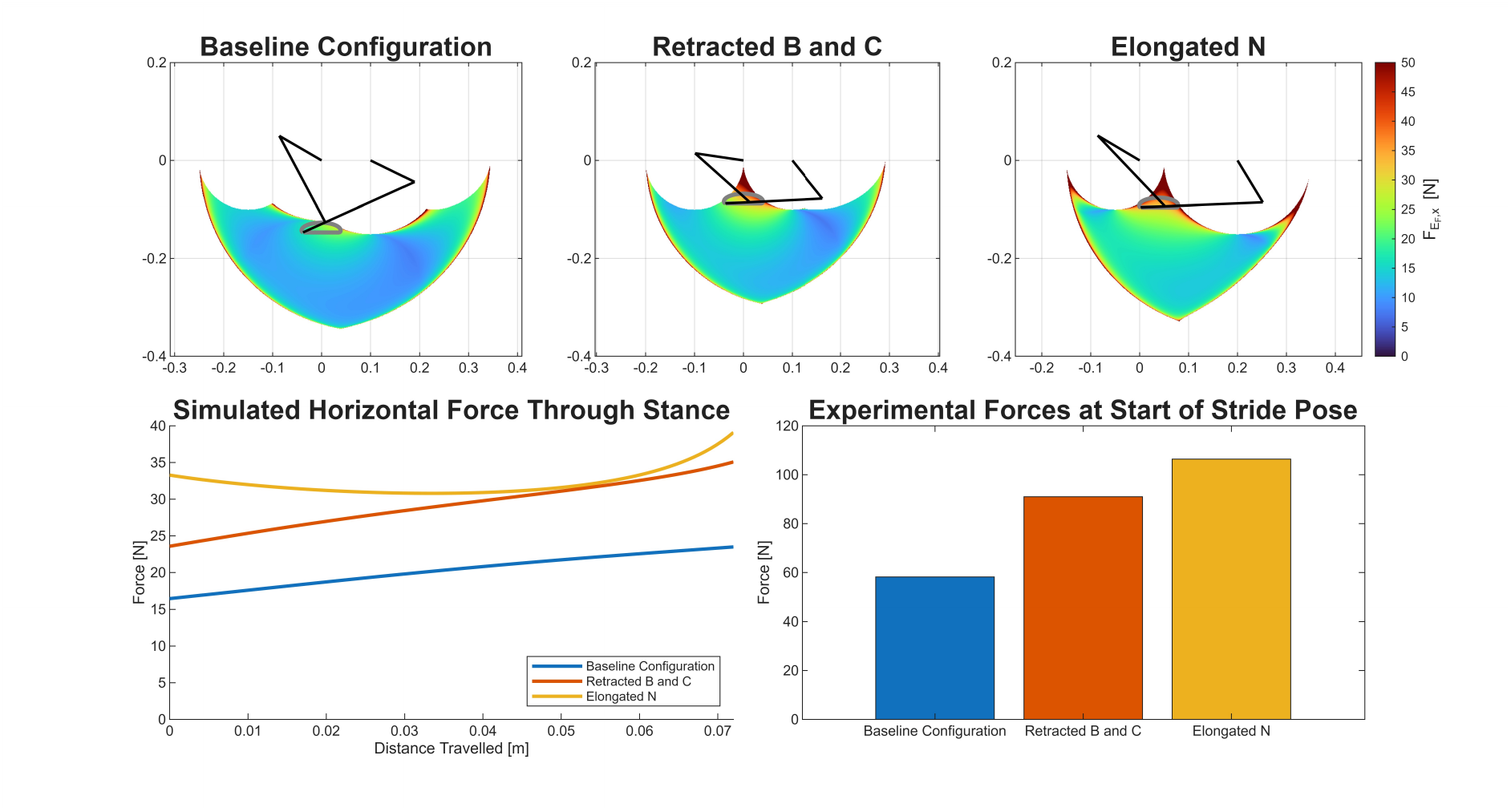}} \caption{Top: Simulations of the placement of the highest average horizontal force profile for the stance portion of a foot path shown in gray. Linkage poses shown in black are the positions at the start of the foot path. Bottom Left: Simulation of the horizontal force $F_{E_F,x}$ through the stance of the walk cycle for each configuration to be tested. Bottom Right: Experimental horizontal forces for each pose are shown in black.} \label{fig:push_test} \end{figure*}

In 'search' mode, the $B$ and $C$ links will be fully extended, and the ground link will be fully retracted to maximize the reachable workspace and lift the chassis of the robot further off the ground. In 'rescue' mode, the $B$ and $C$ links will retract, and the ground link will extend to increase the pushing forces the foot will be able to generate.

To determine suitable locations for the walk cycle while avoiding singularities, the condition number of the Jacobian, $\kappa(J(q))$, was evaluated throughout the workspace of the baseline configuration Figure \ref{fig:condition_num}. A higher condition number indicates that the point is closer to a singularity and that there will be greater error amplification and a lower positioning accuracy \cite{merlet2006jacobian}. The results shown in Figure \ref{fig:condition_num} indicate that the outer regions of the workspace, where the forces are high, correspond to large condition numbers and are therefore unsuitable. In contrast, the central high-force region coincides with lower condition numbers, making it a more favorable area for placing the foot-path trajectory.

\begin{figure*}[!t]
    \vspace{3mm}
    \centering
    \includegraphics[width=1\textwidth,trim =5 250 50 250,clip]{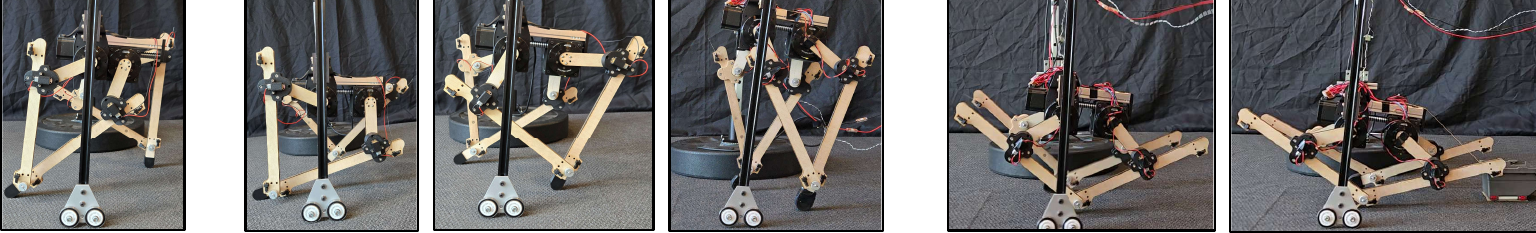}
    \caption{Varying configurations of legs of the bipedal robot for different modes of operation. Left to Right: Retracted $N$ and extended $B$ and $C$ links to maximize workspace. Extended $N$ link with a stepper motor to increase horizontal forces in 3 different poses. Extended $N$ and retracted $B$ and $C$ links with capstan drives to maximize horizontal forces and drag masses. }
    \label{fig:transformation}
\end{figure*}

\vspace{-0.5mm}
\subsection{Real-World Dragging Forces}

To confirm that the length changes would increase the maximum pulling force, the testbed measured the horizontal forces at the start of the stance portion of a sample footpath. The workspace was searched to find where the stance portion of the footpath would have the greatest average pushing force. The path and the poses to be tested are shown in Figure \ref{fig:push_test}. For the tests, an ASANI Mini Crane Scale was used to measure the force the foot could pull the platform back with when the motor bus current was limited to 1 Amp. The peak force generated under static conditions for the baseline, retracted $B$ and $C$ links, and elongated $N$, was tested. The test was repeated 5 times. The results of these tests are shown in Figure \ref{fig:push_test}. The greatest pushing force was achieved when $N$ was elongated, 106 $\pm$ 2 N, which was 48.2 N more than the baseline 58  $\pm$ 1 N. The retracted $B$ and $C$ links exerted a force of 91 $\pm$ 1 N, demonstrating a significant change in force, but not as significant as the change in the $N$ link.

A proof-of-concept bipedal platform suspended on a boom arm, shown in Figure \ref{fig:transformation} was assembled to assess the real-world viability of a dynamically extensible and retractable leg linkage. The system utilized a linear stage with a stepper motor to change the length of the ground link between the motors and capstan drives to change the length of the passive links $B$ and $C$. With this system it is possible to have a leg linkage system with the capability to change the link lengths to adapt to changes in terrain and achieve a more desirable force profile. The system demonstrated an ability to follow a desired foot path and walk, as well as drag at least a 5 lbs. weight when the $B$ and $C$ links were retracted to 18 cm, and the ground link was extended to 13 cm. While this leg did not incorporate a length change of the actively controlled links $A$ and $D$, based on simulation, this actuation would also be able to achieve a desirable change in the force profile, especially in the vertical direction, and the capstan drive designed for the passive links can be adapted to achieve this actuation.

\section{Discussion and Conclusions}

Effectiveness in real-world rescue scenarios for SAR robots requires the ability to drag casualties and step over debris, demanding both terrain adaptability and controlled high-force output. This work proposed and tested a five-bar linkage leg capable of transforming between a "search" mode, a large workspace for rapid traversal, and a "rescue" mode, a high-force output used for dragging in the extraction of casualties. Simulations demonstrate that achievable static forces and the workspace area vary with changes in linkage lengths, highlighting the benefits of an adaptable leg for meeting diverse terrain and force requirements. Experimental results confirm that the leg generates significantly greater pulling force when the ground link $N$ is extended, or the passive links $B$ and $C$ are retracted.

Three criteria define ideal performance during load dragging: a flat, constant-velocity stance phase ($\dot{y_{foot}}=0$, $\dot{x_{foot}}=\text{constant}$); maximum horizontal force without slipping ($F_y=-F_{Normal}$, $F_x=\mu F_y$); and alignment of torque and joint velocity signs ($\tau_A\dot{q_A}>0$, $\tau_D\dot{q_D}>0$). The leg was effectively controlled to meet the first condition. While the horizontal static forces at the foot were demonstrated to be significantly greater after changing the linkage lengths, under dynamic conditions, effects such as slip have not yet been addressed in realistic simulations. During bipedal testing, a slip was observed, reducing both the overall speed of the robot and its dragging capability.

In developing realistic platforms, terrain mechanics, friction, and normal forces at the foot must be considered as part of a successful gait. Additional considerations for scaling include hardware robustness and load-bearing capacity for onboard systems. Overall, the results demonstrate that mechanical reconfiguration offers a promising path toward SAR robots capable of both rapid terrain traversal and effective high-force rescue tasks.
\vspace{-1mm}
\section{Acknowledgments}
\vspace{-1mm}
The authors thank Giovanni Bernal Ramirez, Hwuiyun Park, Elias Smith, Diego Williams, and Edward Lee. 

\vspace{-2.5mm}
\bibliographystyle{ieeetr}
\bibliography{refs}

\end{document}